%% file: paper-arxiv.tex
\documentclass[runningheads]{llncs}
\usepackage{graphicx,subfigure}
\usepackage{amsmath,amssymb} 
\usepackage{color}
\usepackage[width=122mm,left=12mm,paperwidth=146mm,height=193mm,top=12mm,paperheight=217mm]{geometry}

\usepackage[ruled,vlined]{algorithm2e}
\usepackage{mathtools}
\DeclareMathOperator*{\argmin}{argmin}
\DeclareMathOperator*{\argmax}{argmax}

\usepackage[hidelinks=true]{hyperref}
\hypersetup{
  colorlinks   = true, 
  urlcolor     = red, 
  linkcolor    = red, 
  citecolor   = green 
}
\usepackage[capitalize]{cleveref}
\crefname{section}{Sec.}{Section}

\usepackage{booktabs}
\usepackage{tabularx}
\usepackage{multirow}
\usepackage{array}
\newcolumntype{L}[1]{>{\raggedright\let\newline\\\arraybackslash\hspace{0pt}}m{#1}}
\newcolumntype{C}[1]{>{\centering\let\newline\\\arraybackslash\hspace{0pt}}m{#1}}
\newcolumntype{R}[1]{>{\raggedleft\let\newline\\\arraybackslash\hspace{0pt}}m{#1}}

\newcommand{\myparagraph}[1]{\vspace{1em}\noindent\textbf{#1}}

\usepackage{pifont}
\newcommand{\cmark}{\ding{51}}%
\newcommand{\xmark}{\ding{55}}%

\usepackage{caption}
\newcommand{\rot}[2][4.5em]{
  \rotatebox{90}{\parbox{#1}{\raggedright \scriptsize #2}}}

\begin{document}
\pagestyle{headings}
\mainmatter

\title{Joint Optical Flow and Temporally Consistent Semantic Segmentation} 

\titlerunning{Joint Optical Flow and Temporally Consistent Semantic Segmentation}

\authorrunning{Junhwa Hur and Stefan Roth}

\author{Junhwa Hur \and Stefan Roth}


\institute{
	Department of Computer Science, TU Darmstadt\\
}

\maketitle

\begin{abstract}
\input{abstract}
\end{abstract}

\input{introduction}
\input{relatedwork}
\input{approach}
\input{experiments}
\input{conclusion}
\input{acknowledge}

\clearpage

\input{paper-arxiv.bbl}
\end{document}

%% file: abstract.tex

The importance and demands of visual scene understanding have been steadily increasing along with the active development of autonomous systems.
Consequently, there has been a large amount of research dedicated to semantic segmentation and dense motion estimation.
In this paper, we propose a method for jointly estimating optical flow and temporally consistent semantic segmentation, which closely connects these two problem domains and leverages each other.
Semantic segmentation provides information on plausible physical motion to its associated pixels, and accurate pixel-level temporal correspondences enhance the accuracy of semantic segmentation in the temporal domain.
We demonstrate the benefits of our approach on the KITTI benchmark, where we observe performance gains for flow and segmentation.
We achieve state-of-the-art optical flow results, and outperform all published algorithms by a large margin on challenging, but crucial dynamic objects.

%% file: introduction.tex

\section{Introduction}
\label{sec:introduction}

Visual scene understanding from movable platforms has been gaining increased attention due to the active development of autonomous systems and vehicles.
Semantic segmentation and dense motion estimation are two core components for recognizing the surrounding environment and analyzing the motion of entities in the scene.
The performance of techniques in both areas has been steadily increasing, reported and fueled by public benchmarks (e.g.,~KITTI \cite{Geiger:2012:AWR}, MPI Sintel \cite{Butler:2012:NOS}, or Cityscapes \cite{Cordts:2016:CDS}).
Along with the increasing popularity and importance of the two areas, there has been a recent trend in the literature considering how to bridge the two themes and analyzing which benefits these tasks can additionally derive from one another.

There have been a few basic attempts to utilize optical flow to enforce temporal consistency of semantic segmentation in a video sequence \cite{Chen:2011:TCM,Grundmann:2010:EHG,Miksik:2013:ETC}.
Also, segmenting the scene into superpixels (without clear semantics) has been shown to help estimating more accurate optical flow, assuming that object boundaries may give rise to motion boundaries \cite{Sun:2014:LLJ,Yamaguchi:2013:RME,Yamaguchi:2014:EJS}.
Strictly speaking, however, previous work so far simply uses the results from one task as supplementary information for the other, and there have not been many attempts to relate the two tasks more closely or to solve them jointly.
Yet, off-the-shelf motion estimation algorithms are not accurate enough to fully rely on \cite{Chen:2011:TCM,Miksik:2013:ETC}.
The only exception is very recent work that uses both semantic information and segmentation to increase the accuracy of optical flow \cite{Sevilla-Lara:2016:OFS}, however without considering the benefits of temporal correspondence for semantic labeling.

\begin{figure}[t]
\centering     
\includegraphics[width=0.8\textwidth]{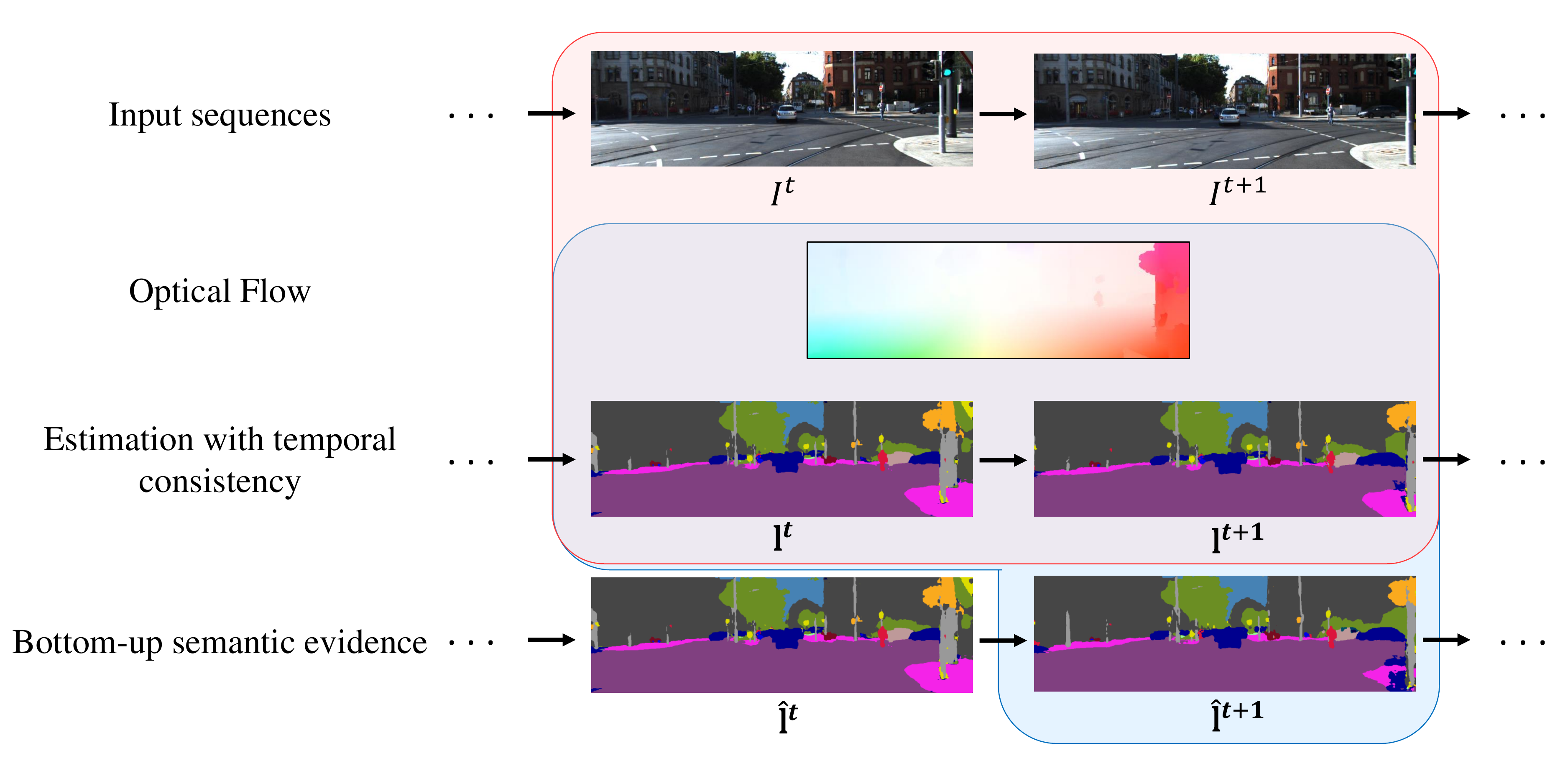}
\vspace{-1.0em}
\caption{\textbf{Overview of our approach.}
The red region contributes to estimating optical flow, and the blue region ensures temporal consistency of the semantic segmentation, both given two frames.
The overlapping region defines the output of our method.}
\label{fig:method}
\vspace{-0.5em}
\end{figure}

In this paper, we address this gap and present an approach for joint optical flow estimation and temporally consistent semantic segmentation from monocular video, in which both tasks leverage each other.
Fig. \ref{fig:method} shows the overview of our method.
We begin by assuming that a bottom-up semantic segmentation for each frame is given.
Then we estimate accurate optical flow fields by exploiting the semantic information from the given semantic segmentation.
The benefit of semantic labels is that they can give us information on the likely physical motion of the associated pixels.
At the same time, accurate pixel-level correspondence between consecutive frames can establish temporally consistent semantic segmentations and help refining the initial results.

We make two major contributions.
First, we introduce an accurate piecewise parametric optical flow formulation, which itself already outperforms the state of the art, particularly on dynamic objects.
Our formulation explicitly handles occlusions to prevent the data term from unduly influencing the results in occlusion areas.
As a result, our method additionally provides occlusion information such as occlusion masks and occlusion types.
Our second contribution is to jointly estimate optical flow and temporally consistent semantic segmentation in a monocular video setting.
For the flow estimation, we additionally apply the epipolar constraint for pixels that should be consistent with the camera ego-motion, as inferred by the semantic information.
At the same time, accurately estimated flow helps to enforce temporal consistency on the semantic segmentation.
We effectively realize these ideas in our joint formulation and make them feasible using inference based on patch-match belief propagation (PMBP) \cite{Besse:2013:PMB}.

Our experiments on the popular KITTI dataset show that our method yields state-of-the-art results for optical flow.
For estimating flows on dynamic foreground objects, which are particularly crucial from an autonomous navigation standpoint, our method outperforms all published optical flow algorithms in the benchmark by a significant margin.




%% file: relatedwork.tex

\section{Related Work}
\label{sec:relatedwork}

\subsubsection{Piecewise parametric flow estimation.}
Piecewise parametric approaches using a homography model have recently shown promising results on standard benchmarks \cite{Butler:2012:NOS,Geiger:2012:AWR} for motion estimation.
Representing the scene as a set of planar surfaces significantly reduces the number of unknowns;
at the same time, parametrizing the motion of surfaces by 9-DoF or 8-DoF transforms ensures sufficient diversity and generality of their motion \cite{Hornacek:2014:HOO,Menze:2015:OSF,Vogel:2014:VC3,Vogel:2013:PRS,Vogel:2015:3SF,Yang:2015:DAO}.
In the stereo setting, Vogel \textit{et al.}~\cite{Vogel:2014:VC3,Vogel:2013:PRS,Vogel:2015:3SF} proposed a scene representation consisting of piecewise 3D planes undergoing 3D rigid motion and demonstrate the most accurate results to date for estimating the 3D scene flow on the KITTI benchmark.

On the other hand, the monocular case with its limited amount of data (i.e.,~two consecutive images) makes the problem more challenging, hence the type of regularization becomes much more important \cite{Hornacek:2014:HOO,Yang:2015:DAO}.
Hornacek \textit{et al.}~\cite{Hornacek:2014:HOO} introduced a 9-DoF plane-induced model for optical flow via continuous optimization.
Their method shows its strength on rigid motions, but is weaker on poorly textured regions because of the lack of global support.
Yang \textit{et al.}~\cite{Yang:2015:DAO} instead use a 8-DoF homography motion in 2D space with adaptive size and shape of the pieces via discrete optimization.

Our approach also relies on an 8-DoF parameterization, which we found to yield accurate optical flow estimates in practice.

\myparagraph{Epipolar constraint-based flow estimation.}
Several approaches have relied on the epipolar constraint for estimating motion  \cite{Baker:2011:DBE}.
Strictly enforcing the constraint gives the benefit of reducing the search space  significantly, but causes an inherent limitation for handling independently moving objects whose motion usually violates the constraint \cite{Kitt:2012:TOF,Mohamed:2015:DOF,Yamaguchi:2013:RME,Yamaguchi:2014:EJS}.
Adding the constraint as a soft prior can resolve this issue, but there is still the challenge of determining where to relinquish the constraint by only depending on the data term \cite{Wedel:2008:DTF,Wedel:2009:SMR}.

Our approach explicitly resolves this ambiguity with the aid of semantic information, which provides information on the physical properties of objects (e.g., static or movable).

\myparagraph{Temporally consistent semantic segmentation.}
Among a broad literature on enabling temporal consistency of video segmentation, we specifically consider the case of semantic segmentation here.
One common way to inject temporal consistency is to utilize motion and structure features from 3D point clouds obtained by Structure from Motion (SfM) \cite{Brostow:2008:SRU,Floros:2012:JTC,Sturgess:2012:CAS}.
Another way is to jointly reconstruct a scene in 3D with semantic labels through a batch process, naturally enabling temporally consistent segmentation \cite{Kundu:2014:JSS,Sengupta:2013:U3S,Zhang:2010:SSU}. In causal approaches that rely on temporal correspondence, previous approaches achieve accurate temporal correspondence using sparse feature tracking \cite{Scharwaechter:2014:SML} or dense flow maps with a similarity function in feature space \cite{Miksik:2013:ETC}. A recent work \cite{Kundu:2016:FSO} introduces feature space optimization for spatio-temporal regularization in partitioned batches with overlaps.

We achieve temporal consistency for semantic segmentation using a jointly estimated, accurate dense flow map, which leverages the semantic information.

\myparagraph{Optical flow with semantics.}
The question of exploiting semantics for optical flow has only received very limited attention so far.
The most related approach is the very recent work by Sevilla-Lara {\em et al.}~\cite{Sevilla-Lara:2016:OFS}, which treats the problem sequentially.
First, the scene is segmented into 3 semantic categories, things, planes and stuff.
Second, motion is estimated individually for these semantic parts and later composited.
In contrast, we treat the entire problem as the minimization of a single unified energy.
Moreover, motion estimation and semantic segmentation are inferred jointly instead of sequentially, hence may mutually leverage each other.
Experimentally, we report significantly more accurate motion estimates for dynamic objects and demonstrate improved segmentation performance.


%% file: approach.tex

\section{Approach}
\label{sec:approach}

The core idea put forward in this paper is that optical flow and semantic segmentation are mutually beneficial and are best estimated jointly to simultaneously improve each other.
Fig.~\ref{fig:method} shows the flow of our proposed method in the temporal domain and explains which elements contribute to achieving which task.
Here, we assume that some initial bottom-up semantic evidence is already given by an off-the-shelf algorithm, such as a CNN (e.g.,~\cite{Long:2015:FCN}), which is subsequently refined by having temporal consistency.
In the red-shaded region, a pair of consecutive images and their refined semantic segmentation contribute to estimating optical flow more accurately.
At the same time, the temporally consistent semantic labeling at time $t+1$ is inferred from its bottom-up evidence, the previously estimated semantic labeling at time $t$, and the estimated flow map.
For longer sequences, our approach proceeds in an online manner on two frames at a time.

Similar to \cite{Yang:2015:DAO}, our formulation is based on an 8-DoF piecewise-parametric model with a superpixelization of the scene.
Superpixels play an important role in our formulation for connecting the two different domains: optical flow and semantic segmentation.
One superpixel represents a global motion as well as a semantic label for its pixels inside, and the motion is constrained by the physical properties that the semantic label implies.
For example, the motion of pixels corresponding to some physically-static objects (e.g., building or road) can only be caused by camera motion.
Thus, enforcing the epipolar constraint on those pixels can effectively regularize their motion.

Another important feature of our formulation is that we explicitly formulate the occlusion relationship between superpixels \cite{Yamaguchi:2013:RME,Yamaguchi:2014:EJS} and infer the occlusion mask as well.
This directly affects the data term such that it prevents occluded pixels from dominating the data term during the optimization.


\subsection{Preprocessing}

\subsubsection{Superpixels.}
As superpixels generally tend to separate objects in images, they can be a good medium for carrying semantic labels and representative motions for their pixels.
Our approach uses the recent state-of-the-art work of Yao {\em et al.}~\cite{Yao:2015:RTC}, which has shown to be well suited for estimating optical flow.

\myparagraph{Semantic segmentation.}
For the bottom-up semantic evidence, we use an off-the-shelf fully convolutional network (FCN) \cite{Long:2015:FCN} trained on the Cityscapes dataset \cite{Cordts:2016:CDS}, which contains typical objects frequent in street scenes.

\myparagraph{Fundamental matrix estimation.}
In order to apply the epipolar constraint on superpixels for which their semantic label tells us that they are surely static objects (e.g. roads, buildings, etc.), our approach requires the fundamental matrix resulting from the camera motion.
We use a standard approach, i.e.~matching SIFT keypoints \cite{Lowe:2004:DIF} and using the 8-point algorithm \cite{Hartley:1997:DEP} with RANSAC \cite{FUNDEST}.

\subsection{Model}
\label{subsec:model}

Our model jointly estimates \emph{(i)} the optical flow between reference frame $I^t$ and the next frame $I^{t+1}$, and \emph{(ii)} a temporally consistent semantic segmentation $\mathbf{l}^{t+1}$ given bottom-up semantic evidence $\mathbf{\hat{l}}^{t+1}$ and the previously estimated semantic labeling $\mathbf{l}^{t}$.
$\mathbf{l}$ is a semantic label probability map, which has the same size as the input image and $L$ channels, where $L$ is the number of semantic classes.
Instead of using a single label, we adopt label probabilities so that we can more naturally and continuously infer the semantic labels in the time domain.
Note that we assume an online setting (i.e.,~no access to future information) and hence infer the segmentation at time $t+1$ rather than $t$.
Optical flow is represented by a set of piecewise motions of superpixels in the reference frame.
We define the motion of a superpixel through a homography and formulate the objective for estimating the 8-DoF homography $\mathbf{H}_s$ of each superpixel $s$ and the temporally consistent semantic segmentation $\mathbf{l}^{t+1}$ as:
\begin{equation}
\begin{split}
E(\mathbf{H}, \mathbf{l}^{t+1}, o, b) = \;& E_\text{D}(\mathbf{H}, o) + \lambda_\text{L} E_\text{L}(\mathbf{H}, \mathbf{l}^{t+1}, o) \\ & + \lambda_\text{P} E_\text{P}(\mathbf{H}) + \lambda_\text{C}  E_\text{C}(\mathbf{H}, o, b) + \lambda_\text{B} E_\text{B}(b).
\label{eq:full-energy}
\end{split}
\end{equation}
Here, $E_\text{D}$, $E_\text{L}$, $E_\text{P}$, $E_\text{C}$, and $E_\text{B}$ denote color data term, label data term, physical constraint term, connectivity term, and boundary occlusion prior, respectively.

We adopt two kinds of occlusion variables: the boundary occlusion label $b$ between two superpixels, and the occlusion mask $o$ defined at the pixel level.
The boundary occlusion label $b$ regularizes the spatial relationship between two neighboring superpixels (i.e.,~co-planar, hinge, left occlusion, or right occlusion) \cite{Yamaguchi:2013:RME,Yamaguchi:2014:EJS}.
The occlusion mask $o$ explicitly models whether a pixel is occluded or not.
One important difference to previous superpixel-based work \cite{Yamaguchi:2014:EJS} is that we additionally infer a pixelwise occlusion mask, which prevents occluded pixels from adversely affecting the data cost.


\myparagraph{Data terms.}
The data terms aggregate photometric differences
\begin{align}
\label{eq:image_data}
E_\text{D}(\mathbf{H}, o) &= \sum_{s \in S} \frac{1}{\lvert s \rvert} \underbrace{ \sum_{\mathbf{p} \in s}  (1 - o_\mathbf{p} ) \rho_\text{D} \Big( I^t(\mathbf{p}), I^{t+1}(\mathbf{p'}) \Big) + o_\mathbf{p} \lambda_\text{o}  }_{\text{image data}}
\end{align}
and semantic label differences
\begin{align}
\label{eq:label_data}
E_\text{L}(\mathbf{H}, \mathbf{l}^{t+1}, o) &= \sum_{s \in S} \frac{1}{\lvert s \rvert} \sum_{\mathbf{p} \in s} \phi_l(\mathbf{H}, \mathbf{l}^{t+1}_\mathbf{p'}, o)\qquad\text{with}\\
\phi_l(\mathbf{H}, \mathbf{l}^{t+1}_\mathbf{p'}, o) &= \frac{1}{2} \sum_{i}^{L} (1 - o_\mathbf{p}) \left\lVert \mathbf{l}^{t+1}_{\mathbf{p'},i} - ( \alpha \mathbf{\hat{l}}^{t+1}_{\mathbf{p'},i} + (1-\alpha)\mathbf{l}^{t}_{\mathbf{p},i}) \right\rVert^2
\end{align}
over each pixel of each superpixel.
Here, $\mathbf{p'}$ is the corresponding pixel in $I^{t+1}$ of pixel $\mathbf{p}$ in $I^{t}$, which is determined according to the homography $\mathbf{H}_\mathcal{S(\mathbf{p})} \in \mathbb{R}^{3 \times 3}$ of its superpixel
\begin{equation}
\label{eq:p_dash}
\mathbf{p'} = \mathbf{H}_\mathcal{S(\mathbf{p})} \mathbf{p},
\end{equation}
where $\mathcal{S}: I^{t} \to S$ is a mapping that assigns a pixel $\mathbf{p}$ to its superpixel $s \in S$.

In the image data term in \cref{eq:image_data}, the function $\rho_D(\cdot,\cdot)$ measures the photometric differences between two pixels using the ternary transform \cite{Stein:2004:ECO} and a truncated linear penalty.
If a pixel $\mathbf{p}$ is occluded (i.e., $o_{\mathbf{p}} = 1$), a constant penalty $\lambda_\text{o}$ is applied.

The label data term in \cref{eq:label_data} measures the distance between two semantic label probability distributions over each pixel: \emph{(i)} our estimation $\mathbf{l}^{t+1}_{\mathbf{p'}}$ and \emph{(ii)} a weighted sum of the previous estimation $\mathbf{l}^{t}_{\mathbf{p}}$, which is propagated by the optical flow, and the bottom-up evidence $\mathbf{\hat{l}}^{t+1}_{\mathbf{p'}}$, while considering its occlusion status.
The motivation of the term is to penalize label differences to the bottom-up evidence and at the same time propagate label evidence over time, except when an occlusion takes place.

\myparagraph{Physical constraint term.}
Semantic labels can provide useful cues for estimating optical flow.
If pixels are labeled as physically static objects, such as building, road, or infrastructure, then they normally do not undergo any 3D motion, hence their observed 2D motion is caused by camera motion and should thus satisfy the epipolar constraint.
We define the corresponding term as
\begin{eqnarray}
  E_\text{P}(\mathbf{H}) &=& \sum_{s \in S} \min ( \phi_{P}(s, \mathbf{H}_s), \lambda_{\text{non\_st}} + \beta [l^t_s \in L_\text{st}] ),\\
  \text{where}\quad\phi_\text{st}(s, \mathbf{H}_s) &=& \frac{1}{\lvert s \rvert} \sum_{\mathbf{p} \in s} {\left\lVert ( \mathbf{p'}^\top \mathbf{F} \mathbf{p} )\right\rVert_1 }
 = \frac{1}{\lvert s \rvert} \sum_{\mathbf{p} \in s} {\left\lVert ( (\mathbf{H}_\mathcal{S(\mathbf{p})} \mathbf{p})^\top \mathbf{F} \mathbf{p} )\right\rVert_1 }
\end{eqnarray}
measures how well the homography matrix $\mathbf{H}_s$ of a superpixel $s$ meets the epipolar constraint from the fundamental matrix $\mathbf{F}$.
For non-static objects, such as pedestrians or vehicles, we still apply the epipolar penalty, however a weak one using a low truncation threshold $\lambda_\text{non\_st}$.
This is motivated by the fact that possibly dynamic objects may in fact stand still and thus obey epipolar geometry, but we do not want to penalize them too much if they do not.
For static objects, on the other hand, we augment the truncation threshold by $\beta$ in order to give a stricter penalty.
$L_\text{st}$ is the set of semantic labels that corresponds to the physically static objects.
$l_s$ is a representative semantic label of superpixel $s$, which has the highest probability over the pixels in the superpixel: $l_s = \argmax_i \sum_{\mathbf{p} \in s} \mathbf{l}^{t}_{\mathbf{p},i}$.

\begin{figure}[t]
\label{fig:occ}
\centering     
\subfigure[]{\label{fig:occ_b}\includegraphics[width=0.61\textwidth]{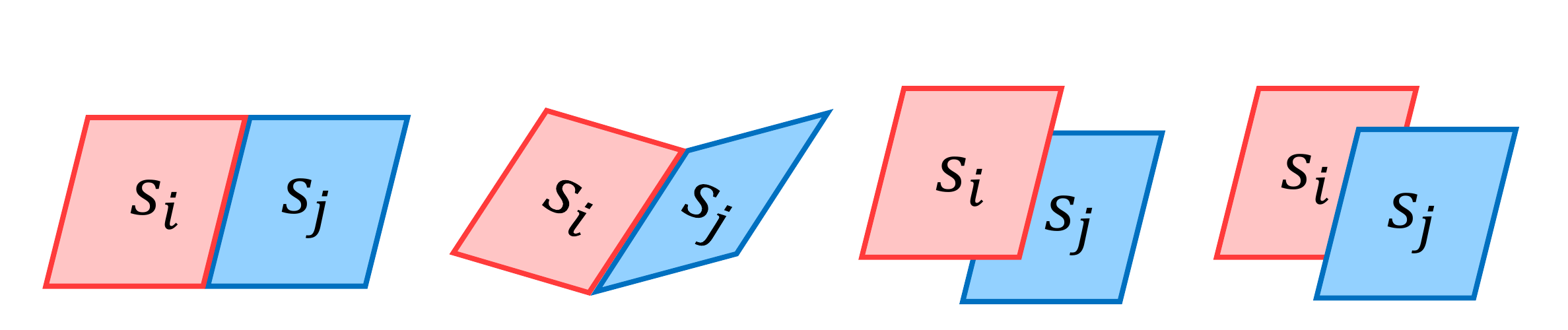}}
\subfigure[]{\label{fig:occ_r}\includegraphics[width=0.38\textwidth]{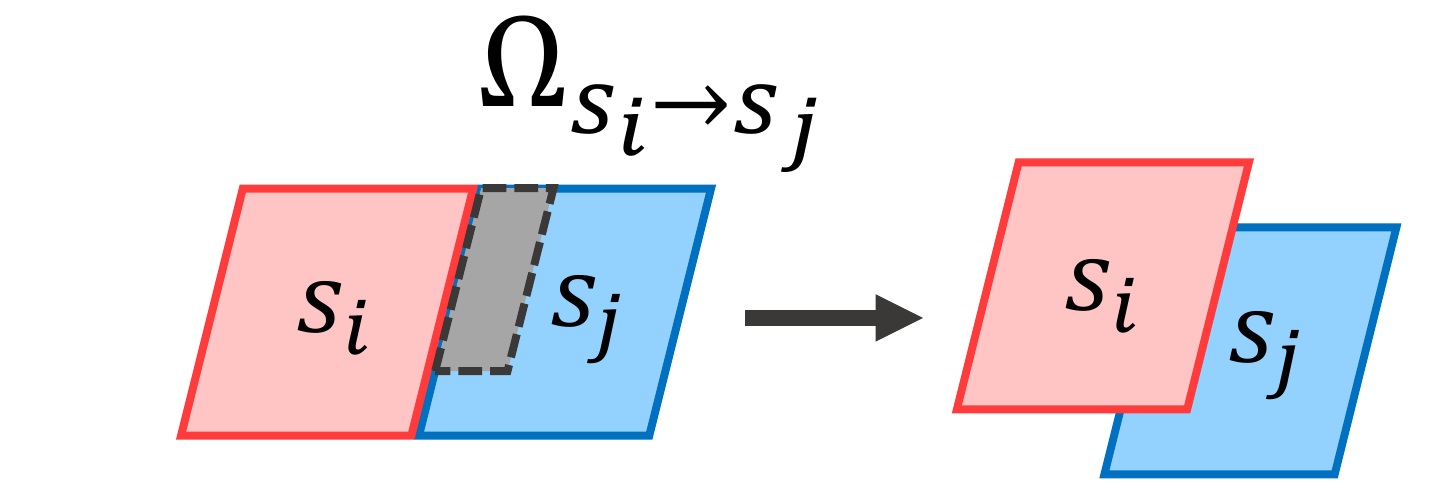}}
\vspace{-1.0em}
\caption{\emph{(a)} Four cases of boundary relations between two superpixels: co-planar, hinge, left occlusion, and right occlusion. \emph{(b)} The visualization of the set of occluded pixels $\Omega_{s_\text{i} \rightarrow s_\text{j}}$ in the case of a left occlusion (black-colored region).}
\vspace{-0.5em}
\end{figure}

\myparagraph{Connectivity term.}
The connectivity term encourages the smoothness of motion between two neighboring superpixels based on their occlusion relationship:
\begin{align}
 E_\text{C}(\mathbf{H}, o, b) &= \sum_{s_i \sim s_j} \phi_\text{C} (\mathbf{H}_{s_i}, \mathbf{H}_{s_j}, o, b_{ij})\\
 \text{with}\quad
 \phi_\text{C} (\mathbf{H}_{s_i}, \mathbf{H}_{s_j}, o, b_{ij}) &= \left\{ \begin{array}{llll}
  \phi_\text{co}(\mathbf{H}_{s_i}, \mathbf{H}_{s_j}, o) &&&  \text{if } b_{ij} = \text{co-planar}, \\
\phi_\text{h}(\mathbf{H}_{s_i}, \mathbf{H}_{s_j}, o) &&&  \text{if } b_{ij} = \text{hinge}, \\
  \phi_\text{occ}(s_i, s_j, o)  &&&  \text{if } b_{ij} = \text{left occlusion}, \\
  \phi_\text{occ}(s_j, s_i, o)  &&&  \text{if } b_{ij} = \text{right occlusion}. \\
 \end{array}\right.
\end{align}
As shown in \cref{fig:occ_b}, the boundary occlusion flag $b_{ij}$ expresses the relationship between two neighboring superpixels $s_i$ and $s_j$ as co-planar, hinge, left-occlusion, or right-occlusion \cite{Yamaguchi:2013:RME,Yamaguchi:2014:EJS}.
This categorization helps to regularize the motion of two superpixels defined by their homography matrices.
We distinguish between three different potentials:
\begin{eqnarray}
\phi_\text{co}(\mathbf{H}_{s_i}, \mathbf{H}_{s_j}, o) &=& \frac{1}{\lvert s_i \cup s_j \rvert } \sum_{\mathbf{p} \in s_i \cup s_j} \lVert \mathbf{H}_{s_{i}} \mathbf{p} - \mathbf{H}_{s_{j}} \mathbf{p} \rVert_1 + \sum_{\mathbf{p} \in s_i \cup s_j} \lambda_\text{imp}[o_p = 1] \\
\phi_\text{h}(\mathbf{H}_{s_i}, \mathbf{H}_{s_j}, o) &=& \frac{1}{ \lvert \mathcal{B}_{s_{i},s_{j}} \rvert } \sum_{\mathbf{p} \in \mathcal{B}_{s_{i},s_{j}}} \lVert \mathbf{H}_{s_{i}} \mathbf{p} - \mathbf{H}_{s_{j}} \mathbf{p} \rVert_1 + \sum_{\mathbf{p} \in s_i \cup s_j} \lambda_\text{imp}[o_p = 1] \\
\phi_\text{occ}(s_\text{f}, s_\text{b}, o) &=&\sum_{\mathbf{p} \in s_\text{f}} \lambda_\text{imp}[o_p = 1] \label{eq:boundaryocc}\\
&&+ \sum_{\mathbf{p} \in s_\text{b}} \Big( \lambda_\text{imp}[ \mathbf{p} \in \Omega_{s_\text{f} \rightarrow s_\text{b}} ][o_p = 0] + \lambda_\text{imp}[ \mathbf{p} \notin \Omega_{s_\text{f} \rightarrow s_\text{b}} ][o_p = 1] \Big) \notag
\end{eqnarray}
These are motivated as follows:
When two superpixels are co-planar, all pixels within should follow the identical homography matrix as they are on the same plane.
For a hinge relationship, only the pixels on the boundary set $\mathcal{B}_{s_{i},s_{j}}$ can satisfy the motion from two superpixels $s_i$ and $s_j$.
In both cases, there should be no occluded pixels, hence we adopt a very large 'impossible' penalty $\lambda_\text{imp}$ to prevent occluded pixels from occurring.
In case that one superpixel occludes another, their motions only affect the occlusion masks.
Eq.~\eqref{eq:boundaryocc} expresses the case that pixels of the front superpixel $s_\text{f}$ occlude some pixels of the back superpixel $s_\text{b}$.
As shown in \cref{fig:occ_r}, $\Omega_{s_\text{f} \rightarrow s_\text{b}}$ is a set of pixels in $s_\text{b}$ that is occluded by some pixels in $s_\text{f}$ from the motion.
All pixels in the front superpixel $s_\text{f}$ should not be occluded, and only pixels in the set of $\Omega_{s_\text{f} \rightarrow s_\text{b}}$ in $s_\text{b}$ should be occluded.

\myparagraph{Boundary occlusion prior.}
Without an additional prior term, the boundary occlusion flag in the connectivity term would prefer to take the occlusion cases.
We thus define a prior term to yield proper biases for each case:
\begin{equation}
 E_\text{B}(b) = \left\{ \begin{array}{llll}
  \lambda_\text{co} [l_{s_i} \neq l_{s_j}] &&&  \text{if } b_{ij} = \text{co-planar}, \\
  \lambda_\text{h} &&&  \text{if } b_{ij} = \text{hinge}, \\
  \lambda_\text{occ}  &&&  \text{if } b_{ij} = \text{occlusion},
 \end{array}\right.
\end{equation}
where $\lambda_\text{occ} > \lambda_\text{h} > \lambda_\text{co} > 0$.
Because it is less likely that two different objects are co-planar in the real world, we only apply the prior penalty for the co-planar case $\lambda_\text{co}$ when the respective semantic labels of the superpixels differ.

\subsection{Optimization}
\label{subsec:optimization}

The minimization of our objective is challenging, as it combines discrete (i.e., $\{l^{t+1}, b, o\}$) and continuous (i.e., $\mathbf{H}$) variables.
We use a block coordinate descent algorithm.
As shown in Alg.~\ref{alg:opt}, we iteratively update each variable in the order:
\emph{(i)} homography matrices $\mathbf{H}$ for superpixels, \emph{(ii)} occlusion variables $b$, $o$, and \emph{(iii)} semantic label probability maps $\mathbf{l}^{t+1}$.
Optimizing the homography matrices $\mathbf{H}$ is especially challenging because the matrices have 8 DoF in 2D space and their parameterization incurs a high-dimensional search space.
We address this using PatchMatch Belief Propagation (PMBP) \cite{Besse:2013:PMB}; see below for details.

Once the motion $\mathbf{H}$ is updated, occlusion variables can be easily updated independently for each pair of neighboring superpixels, while other variables are held fixed.
Given their motions, we first calculate the overlapping region, which can potentially be the occluded region for one of the two superpixels.
Then, we calculate the energy in \cref{eq:full-energy} for all four boundary occlusion cases with the candidate occlusion pixels given.
The boundary occlusion case that has the minimum energy is taken, including the corresponding occlusion mask state.
Finally, the semantic label probability map $\mathbf{l}^{t+1}$ can also be easily updated independently for all superpixels by minimizing label data term in \cref{eq:label_data}.

{
\begin{algorithm}[t]
	\SetAlgoLined
	initialization()\;
	\For{$m = 1$ \text{to n-outer-iters}}{
 		\For{$n = 1$ \text{to n-inner-iters}}{
 			Optimizing $E(\mathbf{H}, \mathbf{l}^{t+1}, o, b)$ for $\mathbf{H}$ using PMBP
	 	}
 		$\{b, o\} = \argmin_{b,o} E(\mathbf{H}, \mathbf{l}^{t+1}, o, b)$\\
	 	$\mathbf{l}^{t+1} = \argmin_{\mathbf{l}^{t+1}} E(\mathbf{H}, \mathbf{l}^{t+1}, o, b)$\\
	}
	\caption{Optimization}
	\label{alg:opt}
\end{algorithm}
}

\myparagraph{Optimizing homography matrices using PMBP.}
Our method optimizes the homography matrices in the continuous domain using PatchMatch Belief Propagation (PMBP) \cite{Hornacek:2014:HOO}.
PMBP is a simple but powerful optimizer based on Belief Propagation.
Instead of using a discrete label set, PMBP uses a set of particles that is randomly sampled and propagated in the continuous domain.
PMBP requires an effective way of proposing the random particles; typically they are obtained from a normal distribution defined over some parameters.
In our approach, however, we devise several strategies for proposing particles of the homography matrices without over-parameterization.
Between two image patches, a superpixel and its corresponding region in the other frame, we estimate the homography matrix by using
\emph{(i)} LK warping, \emph{(ii)} 3 correspondences and the fundamental matrix, \emph{(iii)} 4 randomly perturbed correspondences, and \emph{(iv)} sampled correspondences from neighboring superpixels.
Empirically, we find that these strategies generate reasonable particles without requiring an over-parameterization, and only 5 outer-iterations are enough to be converged.

%% file: experiments.tex

\section{Experiments}
\label{sec:Experiment}

We verify the effectiveness of our approach with a series of experiments on the well-established KITTI benchmark \cite{Geiger:2012:AWR}.
To the best of our knowledge, there is no dataset that simultaneously provides ground truth for optical flow and semantic segmentation in the same scenes; while ground truth for both is available in the KITTI benchmark, the evaluation is carried out on disjoint sequences.

We first evaluate our optical flow results on the KITTI Optical Flow 2015 benchmark and compare to the top-performing algorithms in the benchmark.
In addition, we analyze the effectiveness of the semantics-related terms to understand how effectively the semantic information contributes to the estimation of optical flow.
Finally, we demonstrate  qualitative and quantitative results for temporally consistent semantic segmentation.
We use DiscreteFlow \cite{Menze:2015:DOO} to initialize the flow estimation and utilize the FCN model \cite{Long:2015:FCN} trained on the Cityscapes dataset \cite{Cordts:2016:CDS} for bottom-up semantic segmentation evidence. We set our parameters automatically using Bayesian optimization \cite{Martinez-Cantin:2014:BOB} on the training portion.

\subsection{KITTI 2015 optical flow}

{
\begin{table}[t]
\centering
\footnotesize
\begin{tabular*}{\columnwidth}{@{\extracolsep{\stretch{1}}}*{7}{c}@{}}
\toprule
& \multicolumn{3}{c}{Non-occluded pixels} & \multicolumn{3}{c}{All pixels} \\
\cmidrule(lr){2-4}\cmidrule(l){5-7}
Method	& Fl-bg & Fl-fg & Fl-all & Fl-bg & Fl-fg & Fl-all \\\midrule
MotionSLIC \cite{Yamaguchi:2013:RME}	& $\textbf{6.19 \%}$ & 64.82 \% & 16.83 \% & 14.86 \% & 66.21 \% & 23.40 \% \\
PatchBatch \cite{Gadot:2016:PBB}	& 10.06 \% & 26.21 \% & 12.99 \% & 19.98 \% & 30.24 \% & 21.69 \% \\
DiscreteFlow \cite{Menze:2015:DOO}	& 9.96 \% & 22.17 \% & 12.18 \% & 21.53 \% & 26.68 \% & 22.38 \% \\
SOF \cite{Sevilla-Lara:2016:OFS} 	& 8.11 \% & 23.28 \% & 10.86 \% & $\textbf{14.63 \%}$ & 27.73 \% & $\textbf{16.81 \%}$ \\
$\textbf{Ours (JFS)}$	& 7.85 \% & $\textbf{18.66 \%}$ & $\textbf{9.81 \%}$ & 15.90 \% & $\textbf{22.92 \%}$ & 17.07 \% \\\bottomrule
\end{tabular*}
\smallskip
\caption {\textbf{KITTI Optical Flow 2015:} Comparison to the published top-performing optical flow methods in the benchmark.
Our method leads to state-of-the-art results and significantly increases the performance on challenging dynamic regions (\emph{fg}).}
\label{table:flow2015}
\vspace{-1.5em}
\end{table}
}

{
\begin{figure}[t]
	\centering
		\includegraphics[width=0.32\linewidth]{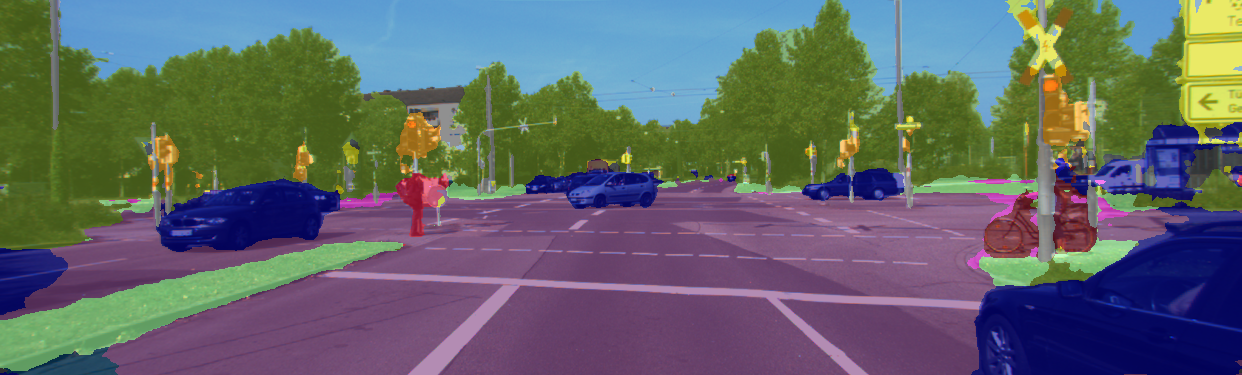}
		\includegraphics[width=0.32\linewidth]{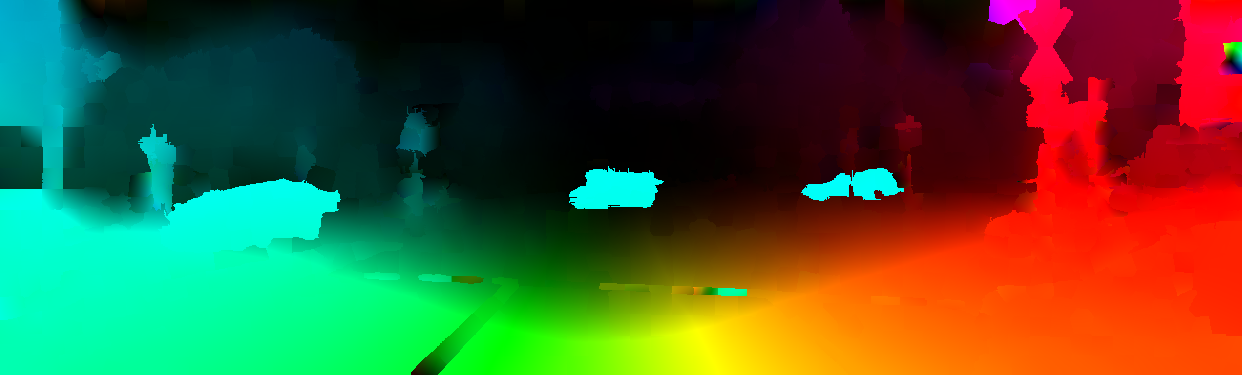}
		\includegraphics[width=0.32\linewidth]{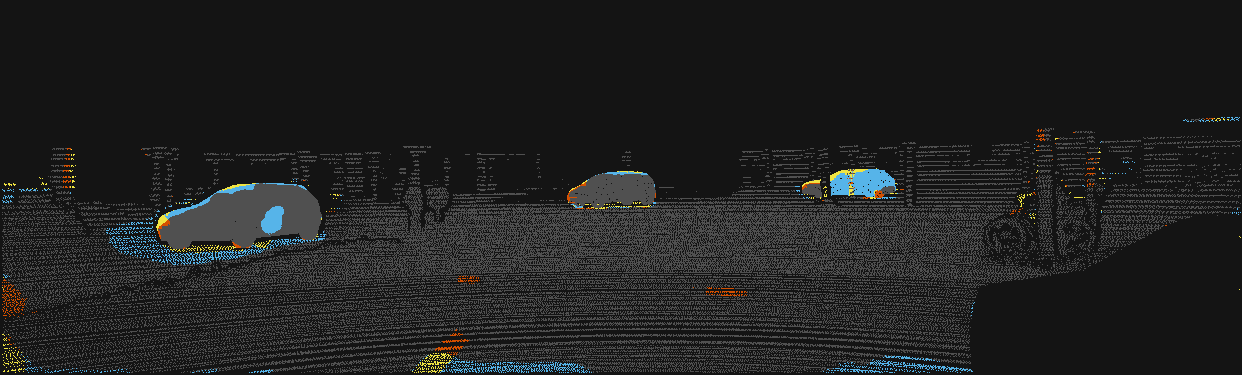} \\

		\includegraphics[width=0.32\linewidth]{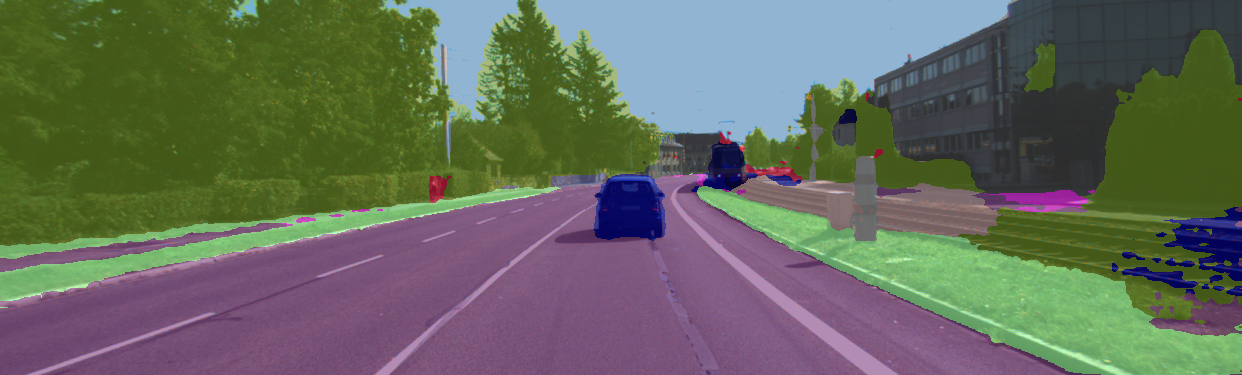}
		\includegraphics[width=0.32\linewidth]{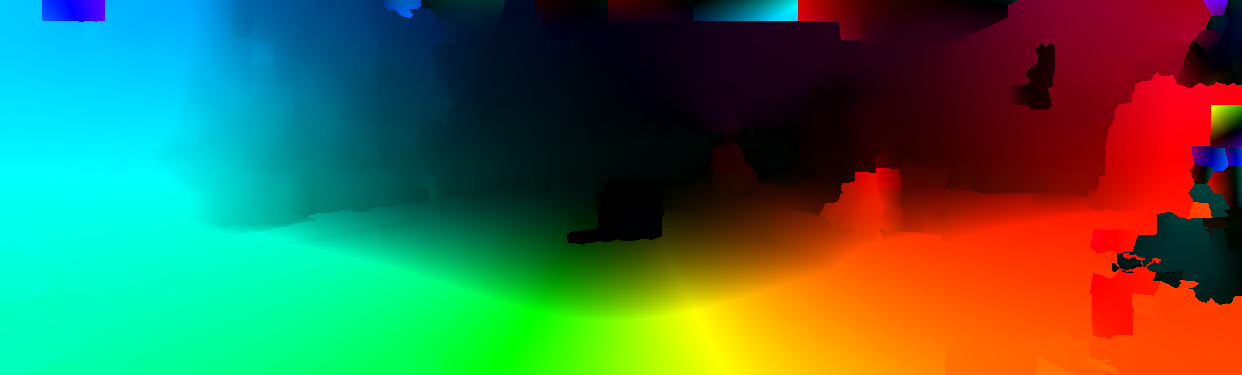}
		\includegraphics[width=0.32\linewidth]{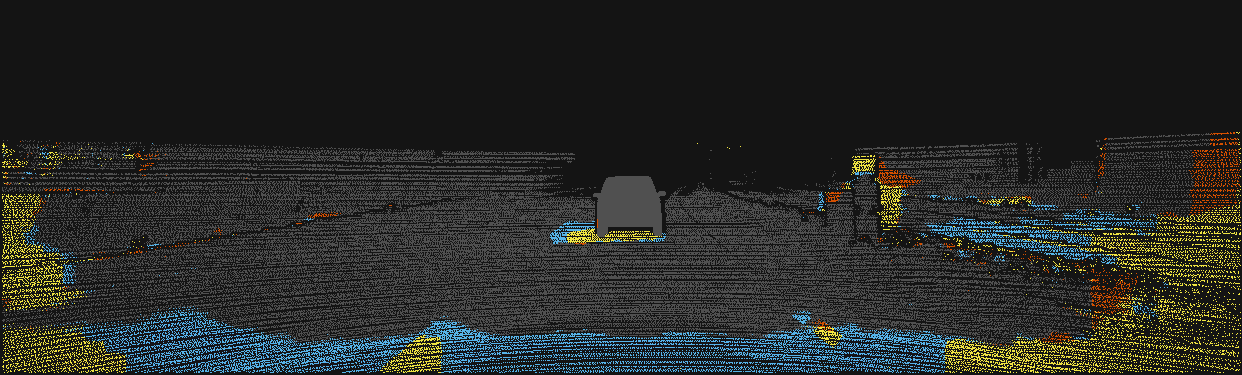} \\

		\includegraphics[width=0.32\linewidth]{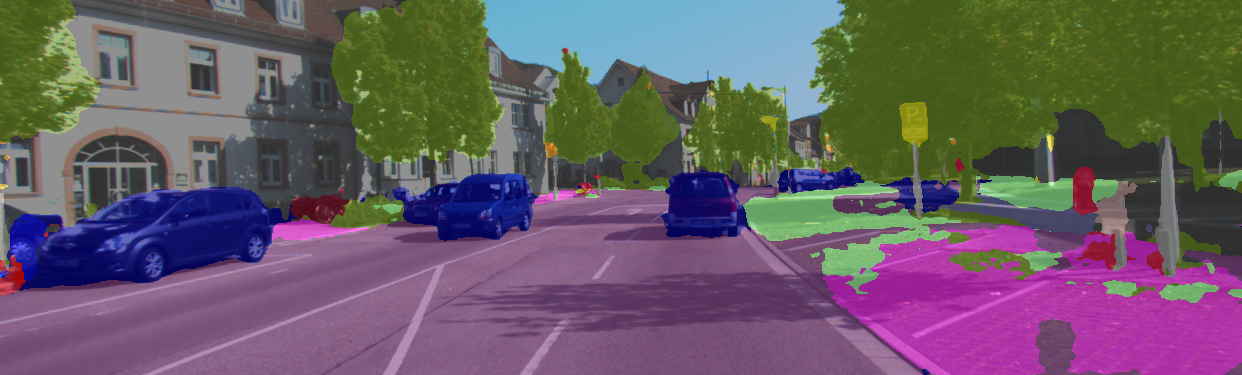}
		\includegraphics[width=0.32\linewidth]{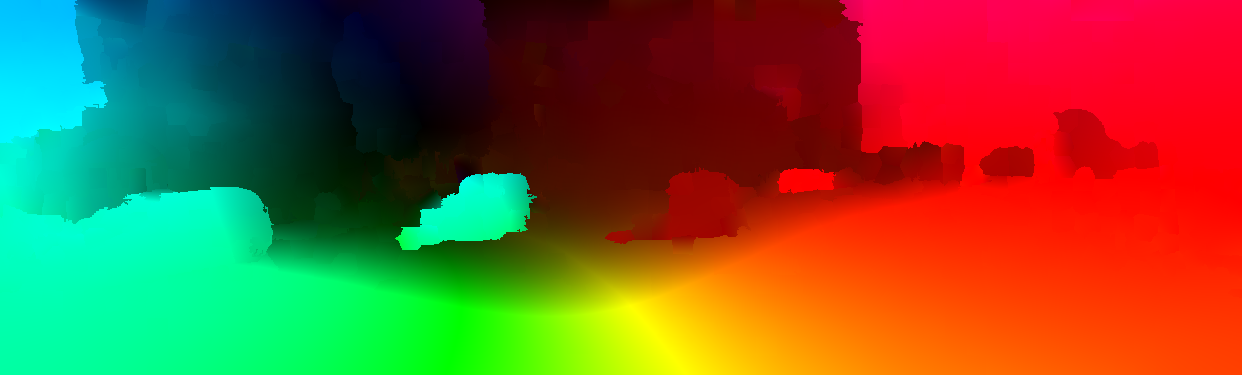}
		\includegraphics[width=0.32\linewidth]{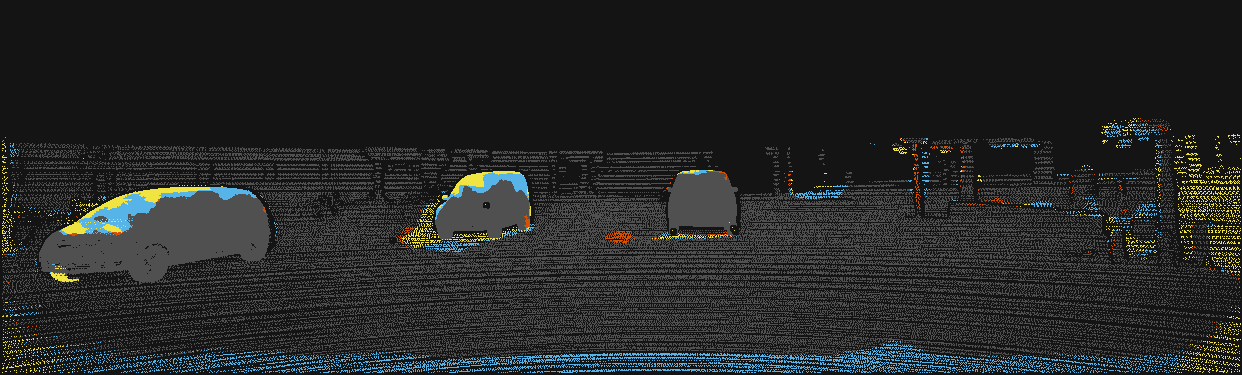} \\

		\includegraphics[width=0.32\linewidth]{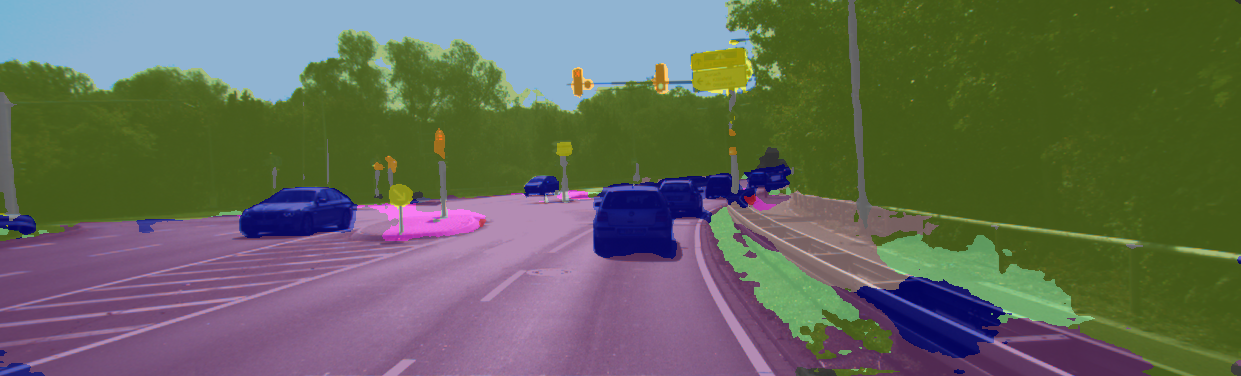}
		\includegraphics[width=0.32\linewidth]{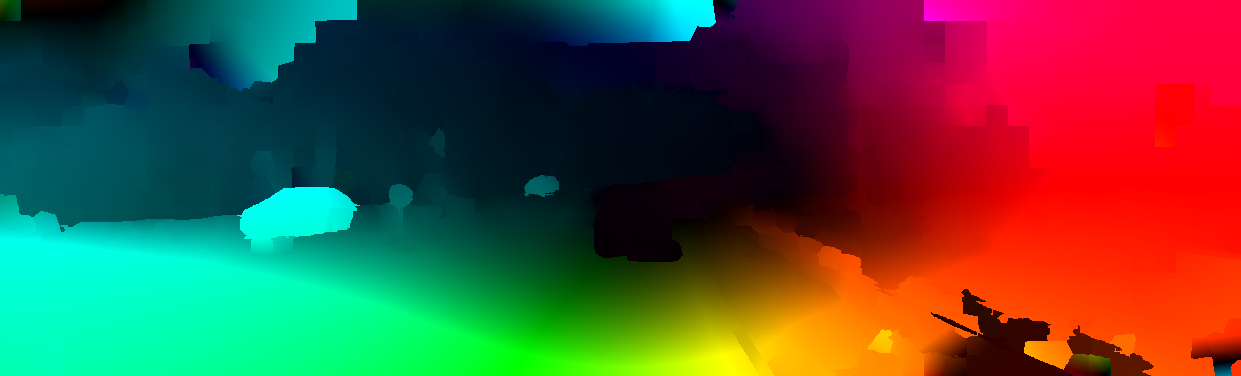}
		\includegraphics[width=0.32\linewidth]{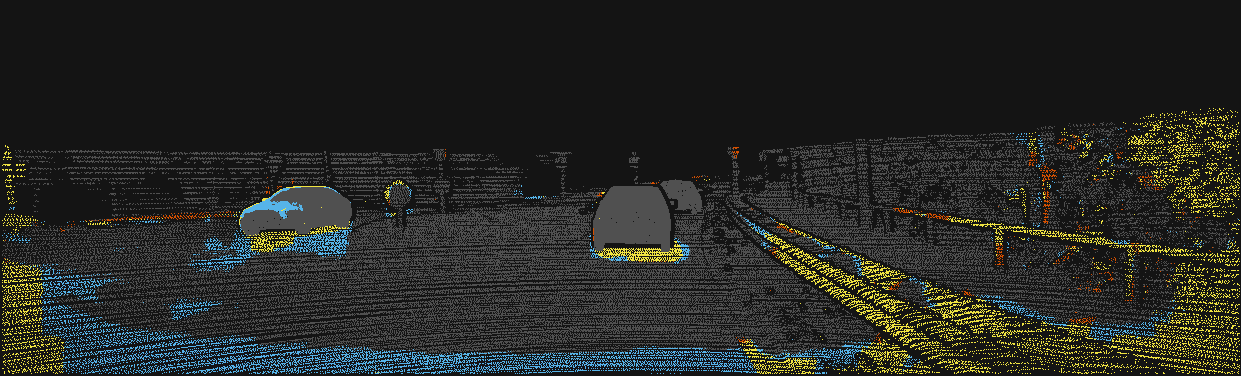}\\[0em]
		\vspace{-0.5em}
\caption{ \textbf{Results on KITTI Optical Flow 2015}. \textbf{Left}: Source images overlaid with semantic segmentation results. \textbf{Middle}: Our flow estimation results. \textbf{Right}: Qualitative comparison with DiscreteFlow: gray pixels -- both methods correct, skyblue pixels -- our method is correct but DiscreteFlow is not, red pixels -- DiscreteFlow is correct but ours is not, and yellow pixels -- both failed.}
\label{fig:exp_flow}
\vspace{-1.0em}
\end{figure}
}

We compare to the top-scoring optical flow methods on the KITTI Optical Flow 2015 benchmark, which have been published at the time of submission.
Note that we do not consider scene flow methods here, as they have access to multiple views.
\Cref{table:flow2015} shows the results.
\emph{Fl-bg}, \emph{Fl-fg}, and \emph{Fl-all} denote the flow error evaluated for background pixels only, foreground pixels only, or for all pixels, respectively.
Our method outperforms all top-scoring methods when considering all non-occluded pixels and performs very close to the leading method when considering all pixels.
Especially for the flow of dynamic foreground objects, our method outperforms all published results by a large margin.
This is of particular importance in the domain of autonomous navigation where understanding the motion of other traffic participants is crucial.
This substantial performance gain stems from several design decisions.
First, our piecewise motion representation effectively abstracts the planar surfaces of foreground vehicles, and the 8-DoF homography successfully describes the rigid motion of each surface.

The soft epipolar constraint of our model, derived from the jointly estimated semantics, contributes to the flow estimation particularly on background pixels and clear performance gains are observed for non-occluded pixels.
When including occluded pixels, however, SOF \cite{Sevilla-Lara:2016:OFS} slightly outperforms ours.
The main reason is that their localized layer approach and planar approximation with large pieces can regularize the occluded regions better than our piecewise model based on superpixels.
In future work, this gap may be addressed through an additional global support model or coarse-to-fine estimation.
MotionSLIC \cite{Yamaguchi:2013:RME} still performs better than ours on background pixels by strictly enforcing the epipolar constraint. As a trade-off, however, their strict epipolar constraint yields significant flow errors for foreground pixels and eventually increases the overall error.

\Cref{fig:exp_flow} shows visual results on the KITTI dataset (visualized as in \cite{Sevilla-Lara:2016:OFS}) and provides a direct comparison to DiscreteFlow, which highlights where the performance gain over the initialization originates.
Our method provides more accurate flow estimates on foreground objects, but also on static objects.

\subsection{Effectiveness of semantic-related terms}

Next we analyze the effectiveness of the semantic-related terms,  the epipolar constraint term and the label data term, in order to understand how much the semantic information contributes to optical flow estimation over our basic piecewise optical flow model.
We turned off each term and evaluated how each setting affects the flow estimation results on the KITTI Flow 2015 training dataset.
The analysis is shown in \cref{table:flow2015_term}.

We find that the label term clearly contributes to more accurate flow estimation overall, but it has a side-effect on background areas where the initial semantic segmentation may have some outliers.
Using the epipolar constraint term results in more accurate flow estimates on background areas, which majorly satisfy the epipolar assumption.
On foreground objects, however, the flow error slightly increases.
This performance loss is coming from the trade-off of our assumption that non-static objects (e.g.,~vehicles) sometimes do not move, which made us apply the epipolar cost but with a small truncation threshold.


One interesting observation is that our basic piecewise flow model, without the semantic-related terms, still demonstrates competitive performance for estimating optical flow on non-occluded pixels.

{
\begin{table}[t]
\centering
\footnotesize
\begin{tabular*}{\columnwidth}{@{\extracolsep{\stretch{1}}}*{8}{c}@{}}
	\toprule
	\multicolumn{2}{c}{Usage of terms} & \multicolumn{3}{c}{Non-occluded pixels} & \multicolumn{3}{c}{All pixels} \\
	\cmidrule(lr){1-2}\cmidrule(lr){3-5}\cmidrule(lr){6-8}
	Label & Epi & Fl-bg & Fl-fg & Fl-all & Fl-bg & Fl-fg & Fl-all \\\midrule
	\cmark & \cmark	& 8.27 \% & 17.40 \% & $\textbf{9.83 \%}$ & 16.44 \% & 20.02 \% & $\textbf{16.98}$ \% \\
	\cmark & \xmark	& 8.45 \% & $\textbf{16.97 \%}$ & 9.90 \% & 16.73 \% & $\textbf{19.61 \%}$ & 17.17 \% \\
	\xmark & \cmark	& $\textbf{8.20 \%}$ & 17.82 \% & 9.84 \% & $\textbf{16.35 \%}$ & 20.41 \% & 16.99 \% \\
	\xmark & \xmark & 8.51 \% & 17.21 \% & 10.00 \% & 16.84 \% & 19.86 \% & 17.31 \% \\\bottomrule
\end{tabular*}
\smallskip
\caption {\textbf{Effectiveness of semantic-related terms:}
The performance of our basic piecewise optical flow model is boosted further (KITTI 2015 training set).}
\label{table:flow2015_term}
\vspace{-1.0em}
\end{table}
}

\subsection{Temporally consistent semantic segmentation}
We finally evaluate the performance of our temporally consistent semantic segmentation on a sequence from the KITTI dataset, which has a 3rd-party ground truth semantic annotation \cite{Ros:2015:VBO}.
This, however, is a preliminary result, since the semantic segmentation model we used here is trained on the higher-resolution Cityscapes dataset \cite{Cordts:2016:CDS}, which possesses somewhat different statistics.
Better results are expected from a custom-trained model.
Table~\ref{table:sem._seg.} shows that our joint approach increases the segmentation accuracy over the bottom-up segmentation results \cite{Long:2015:FCN} by 2 percentage points in the intersection-over-union (IoU) metric.
The accuracy is increased on all object classes except for the pole class, which is not well captured by our superpixels.
\cref{fig:sem._seg._frames} shows our results on three consecutive frames, and \cref{fig:sem._seg._comparison} demonstrates our performance gain/loss over the bottom-up segmentation using the visualization of \cref{fig:exp_flow}.
With the aid of accurate temporal correspondences, our method revises inconsistent results and effectively reduces false positives in the time domain.

{
\begin{table}[t]
\centering
\footnotesize
\begin{tabular*}{\columnwidth}{@{\extracolsep{\stretch{1}}}*{13}{c}@{}}
	\toprule
	IoU (\%) & \rot{sky} & \rot{building} & \rot{road} & \rot{sidewalk} & \rot{fence} & \rot{vegetation} & \rot{pole} & \rot{car} & \rot{sign} & \rot{pedestrian} & \rot{cyclist} & \rot{$\textbf{mean}$} \\\midrule
	FCN \cite{Long:2015:FCN} & 69.35 & 78.53 & 73.75 & 38.19 & 33.33 & 68.37 & 23.68 & 77.60 & 31.27 & 20.11 & 21.42 & 48.69\\
	Ours & 71.80 & 79.97 & 77.99 & 41.01 & 36.27 & 69.21 & 16.44 & 78.58 & 39.05 & 23.50 & 25.44 & 50.84 \\\bottomrule
\end{tabular*}
\smallskip
\caption {\textbf{Performance of temporally consistent semantic segmentation}.}
\label{table:sem._seg.}
\vspace{-1.0em}
\end{table}
}

{
\begin{figure}[t]
\centering
\subfigure[Results on three consecutive frames.]{
	\centering
	\setlength\tabcolsep{0.3pt}
	\renewcommand{\arraystretch}{0.2}
	\begin{tabular}{c >{\centering\arraybackslash}m{.29\textwidth} >{\centering\arraybackslash}m{.29\textwidth} >{\centering\arraybackslash}m{.29\textwidth}}
		{\scriptsize FCN \cite{Long:2015:FCN}} &
		\includegraphics[width=\linewidth]{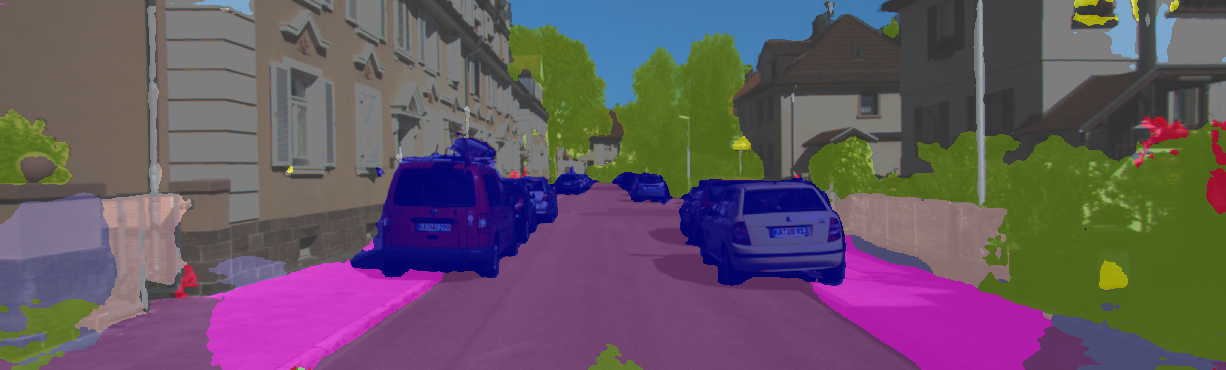}&
		\includegraphics[width=\linewidth]{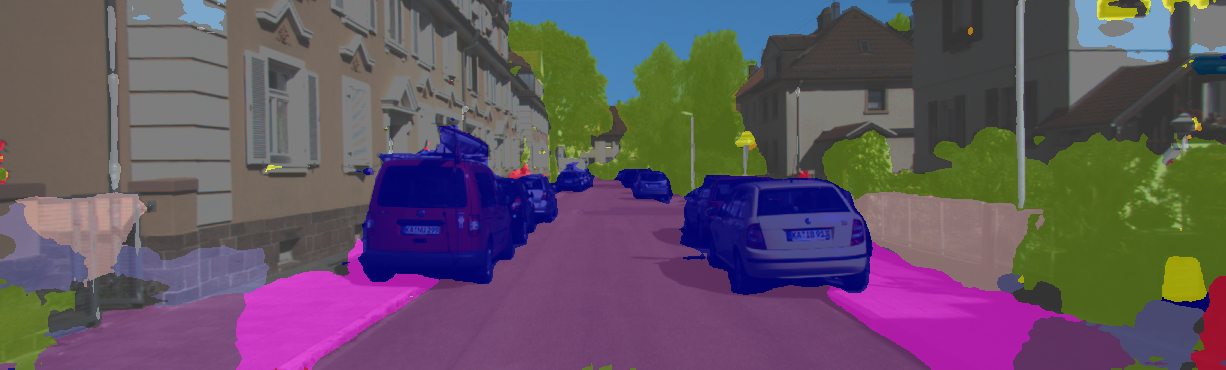}&
		\includegraphics[width=\linewidth]{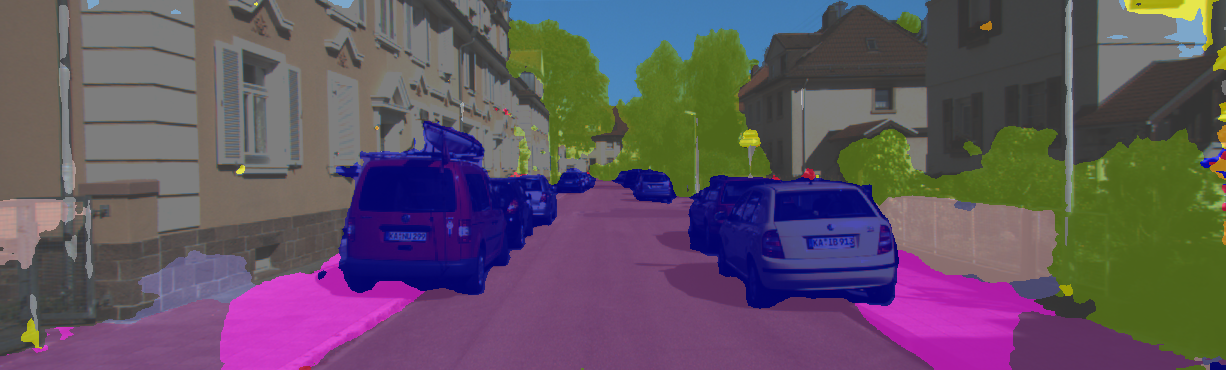} \\
		{\scriptsize Ours} &
		\includegraphics[width=\linewidth]{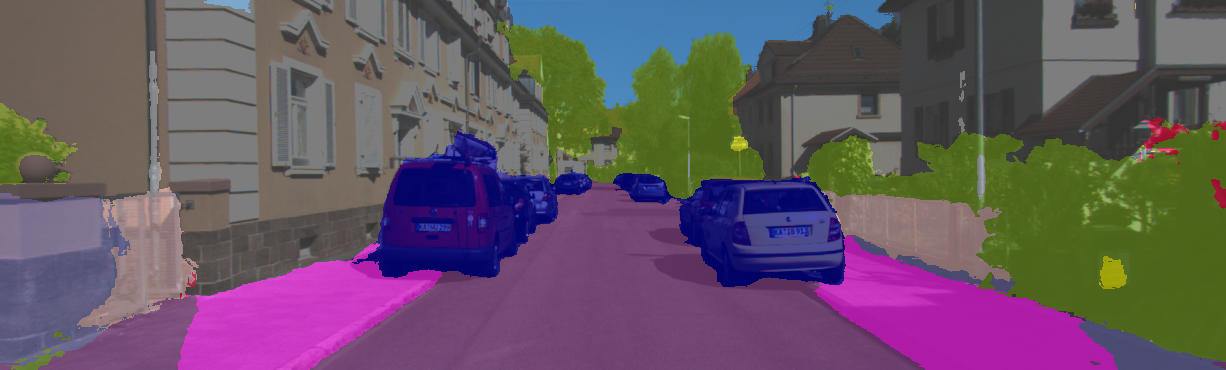}&
		\includegraphics[width=\linewidth]{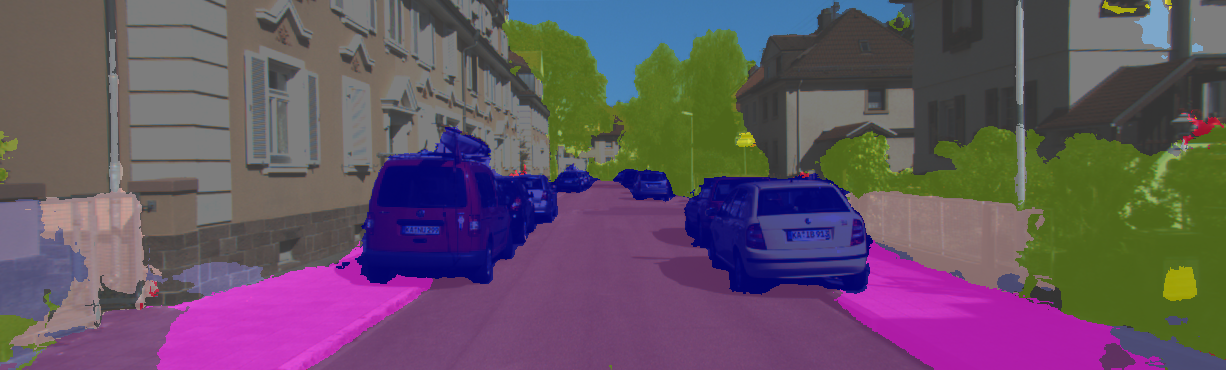}&
		\includegraphics[width=\linewidth]{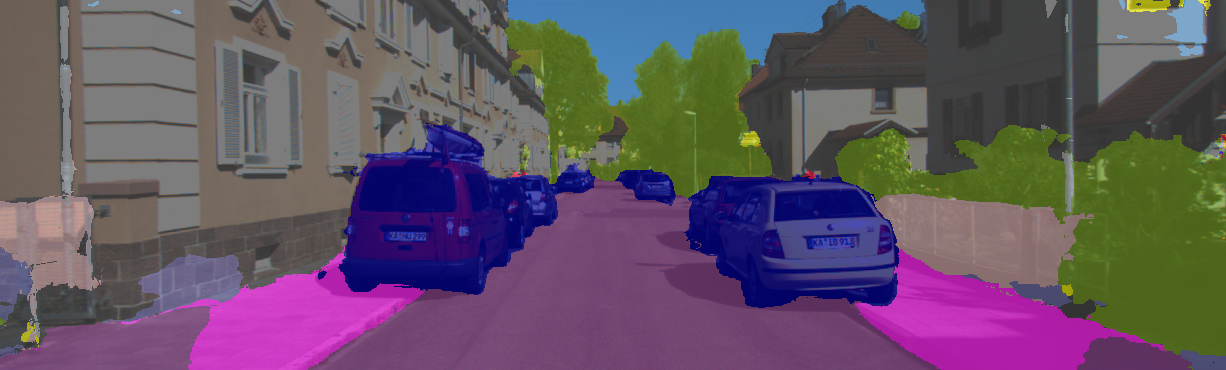} \\
		& time t-1 & time t & time t+1\\[0.3em]
	\end{tabular}
	\label{fig:sem._seg._frames}
}

\subfigure[Performance gain/loss over bottom-up semantic segmentation.]{
	\centering
	\hspace{1cm}
	\includegraphics[width=0.56\textwidth]{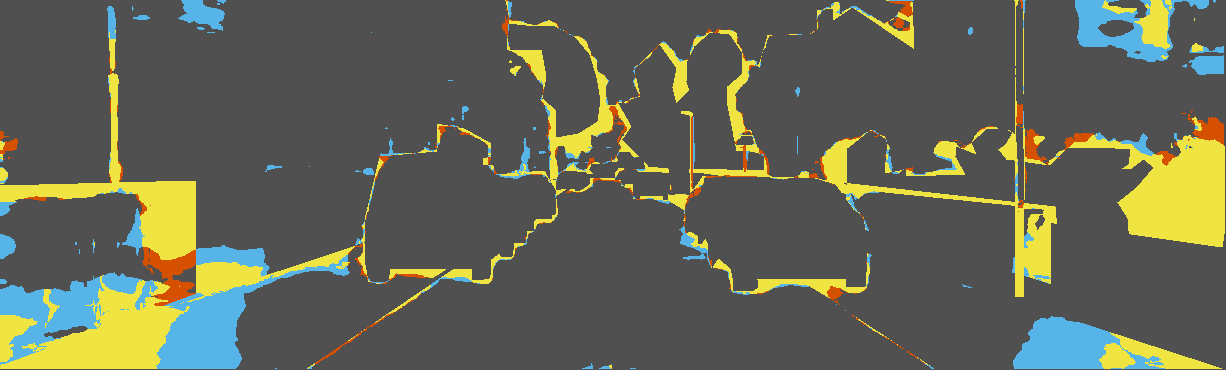}
	\hspace{1cm}
	\label{fig:sem._seg._comparison}
}
\vspace{-1.0em}
\caption{\textbf{Temporally consistent semantic segmentation results}.}
\label{fig:sem._seg.}
\end{figure}
}

%% file: conclusion.tex

\section{Conclusion}
\label{sec:Conclusion}
We have proposed a method for jointly estimating optical flow and temporally consistent semantic segmentation from monocular video.
Our results on the challenging KITTI benchmark demonstrated that both tasks can successfully leverage each other.
A piecewise optical flow model with PMBP inference builds the basis and itself already achieves competitive results.
Embedding semantic information through label consistency and epipolar constraints further boosts the performance.
For dynamic objects, which are particularly important from the viewpoint of autonomous navigation, our method outperforms all published results in the benchmark by a large margin.
Preliminary results on temporally consistent semantic segmentation further demonstrate the benefit of our approach by reducing false positives and  flickering.
We believe that a refinement of the superpixels may lead to further performance gains in the future.

%% file: acknowledge.tex

\small{\myparagraph{Acknowledgement.}
We thank Marius Cordts for providing a pre-trained semantic segmentation model. The research leading to these results has received funding from the European Research Council under the
European Union's Seventh Framework Programme (FP7/2007--2013) / ERC
Grant Agreement No.~307942.}%